\documentclass[runningheads]{llncs}
\usepackage[T1]{fontenc}
\usepackage{orcidlink}
\usepackage{graphicx}
\usepackage{subcaption}
\usepackage{tcolorbox}
\usepackage{listings}
\usepackage[ruled,vlined,linesnumbered]{algorithm2e}
\usepackage{amsmath}
\usepackage{booktabs} 
\usepackage{tabularx}
\usepackage{float}
\usepackage{subcaption} 

\lstdefinestyle{mystyle}{
  basicstyle=\ttfamily\small,
  breaklines=true,
  frame=none,
  numbers=none,
  xleftmargin=1em,
  aboveskip=0.5em,
  belowskip=0.5em
}

\begin{document}
\title{SGTA: Scene-Graph Based Multi-Modal Traffic Agent for Video Understanding}
%
%
\author{Xingcheng Zhou\inst{1}\orcidlink{0000-0003-1178-5221} \and
Mingyu Liu\inst{1}\orcidlink{0000-0002-8752-7950} \and
Walter Zimmer \inst{1}\orcidlink{0000-0003-4565-1272} \and
Jiajie Zhang\inst{1}\orcidlink{0009-0000-3485-8265} \and
Alois Knoll\inst{1}\orcidlink{0000-0003-4840-076X}}
%
%
\institute{Technical University of Munich, Chair of  Robotics, Artificial Intelligence (AI) and Embedded Systems, Boltzmannstraße 3, 85748 Garching bei München, Germany
\email{xingcheng.zhou@tum.de,mingyu.liu@tum.de,walter.zimmer@tum.de, \\jiajie.zhang@tum.de, knoll@in.tum.de}\\
}
\maketitle              
\begin{abstract}
We present \textbf{S}cene-\textbf{G}raph Based Multi-Modal \textbf{T}raffic \textbf{A}gent (SGTA), a modular framework for traffic video understanding that combines structured scene graphs with multi-modal reasoning. It constructs a traffic scene graph from roadside videos using detection, tracking, and lane extraction, followed by tool-based reasoning over both symbolic graph queries and visual inputs. SGTA adopts ReAct to process interleaved reasoning traces from large language models with tool invocations, enabling interpretable decision making for complex video questions. Experiments on selected TUMTraffic VideoQA dataset sample demonstrate that SGTA achieves competitive accuracy across multiple question types while providing transparent reasoning steps. These results highlight the potential of integrating structured scene representations with multi-modal agents for traffic video understanding.

\end{abstract}

\section{Introduction}

Understanding complex traffic scenes from roadside videos is essential for intelligent transportation systems, traffic safety monitoring, and autonomous driving. Recent advances in large language models (LLMs), vision–language models (VLMs), and multi-modal agent have enabled visual question answering on traffic scenraios \cite{wang2024videoagentlongformvideounderstanding,fan2024videoagentmemoryaugmentedmultimodalagent,xie2024largemultimodalagentssurvey}. However, existing methods often lack interpretability and struggle with fine-grained spatio-temporal reasoning, especially in traffic scenarios \cite{xie2024largemultimodalagentssurvey}. To address these limitations, we propose the \textbf{Scene-Graph Based Multi-Modal Traffic Agent (SGTA)} for traffic video understanding. SGTA first constructs a traffic scene graph \cite{Chang_2023} that encodes spatio–temporal relations among objects, lanes, and frames using detection, tracking, and map priors. Based on this structured representation, SGTA employs a ReAct-style \cite{yao2023reactsynergizingreasoningacting} reasoning framework that interleaves reasoning traces from an LLM with tool calls over the scene graph and visual inputs. This design allows SGTA to decompose complex queries into sequences of interpretable reasoning steps while integrating both symbolic and visual information.  

\section{Related Work}
\vspace{-8mm}
\begin{figure}[H]
    \centering
    \begin{minipage}[t]{0.48\linewidth}
        \centering
        \includegraphics[width=\linewidth]{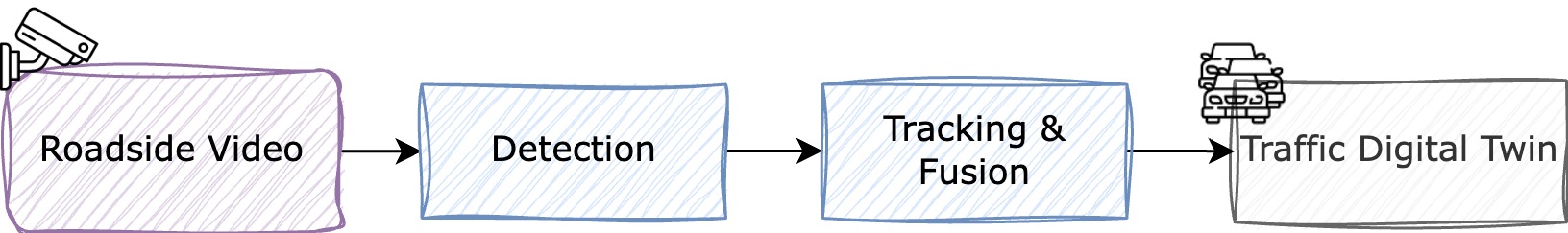}
        \\(a) Traffic digital twin approach.
    \end{minipage}
    \hfill
    \begin{minipage}[t]{0.48\linewidth}
        \centering
        \includegraphics[width=\linewidth]{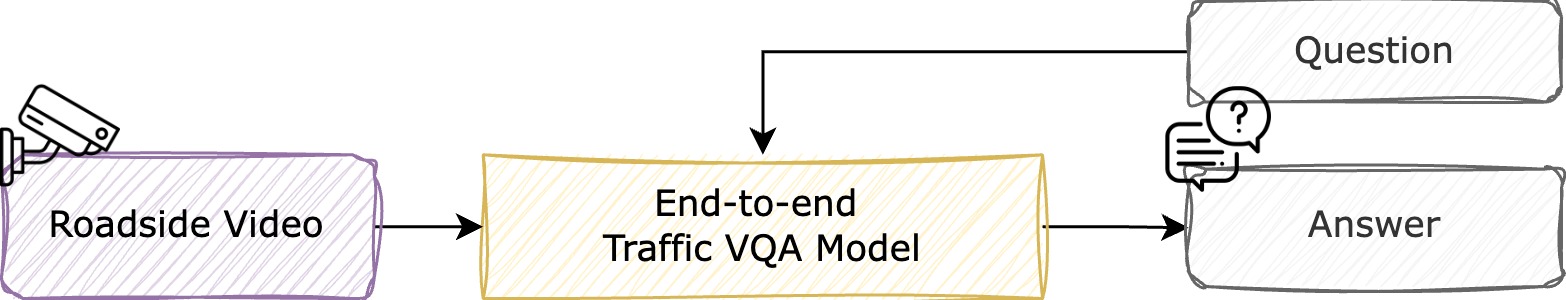}
        \\(b) End-to-end traffic VQA models. 
    \end{minipage}

    \vspace{3mm} 
    \includegraphics[width=0.48\linewidth]{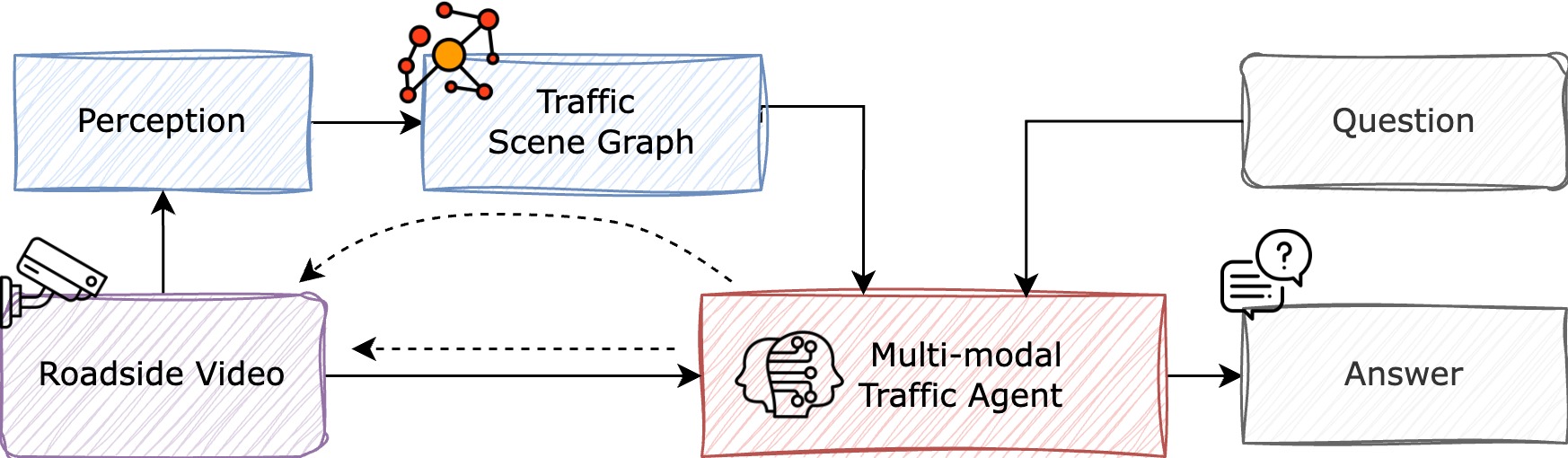}
    \\(c) Scene-graph based multi-modal traffic agent (ours).

    \caption{Comparison of paradigms for traffic scene understanding.}
    \label{fig:comparison}
\end{figure}
\vspace{-5mm}


\subsection{Roadside Traffic Digital Twin}
Roadside traffic digital twin is a digital representation of the traffic state that mirrors the physical environment in real time. Building such a system normally requires roadside sensing with cameras, lidars, and radars, calibration and synchronization across sites, and multi-sensor fusion that outputs object position within unified coordinate system. Networking and edge-cloud computing are integrated to satisfy bandwidth and latency requirements. As a pioneer system, Providentia~\cite{krämmer2021providentialargescalesensor} demonstrated the feasibility of highway-scale traffic digital twins on the A9 test field and later extended the concept to urban intersections using distributed architectures with multi-modal sensor fusion. The traffic digital twin system is shown in Fig \ref{fig:comparison}-(a). Autotech.agil project \cite{van2023autotech} provides modular, service-oriented architectures beyond vehicles to the entire traffic system to enable scalable and updatable digital twin frameworks for cooperative mobility. For roadside perception algorithms, BEVHeight~\cite{yang2023bevheightrobustframeworkvisionbased} generates BEV features from roadside images to construct 3D roadside detection, and WARM-3D~\cite{zhou2024warm3dweaklysupervisedsim2realdomain} introduces a semi-supervised domain adaptation framework to improve the real-world 3D roadside perception ability. Recent datasets such as TUMTraf series~\cite{TUMTrafV2X}, and DAIR-V2X~\cite{yu2022dairv2xlargescaledatasetvehicleinfrastructure} provide large-scale roadside data that benchmark the perception ability of roadside traffic digital rwin systems.

\subsection{Traffic Scene Understanding with Language Models}
With the success of LLMs and VLMs, they have been increasingly adopted for traffic scene understanding, aiming to interpret visual observations and answer natural language queries in complex road environments. VLMs directly process multi-modal inputs and provide end-to-end scene interpretation, as shown in Fig. \ref{fig:comparison}-(b). Zhou \cite{zhou2024gpt4vtrafficassistantindepth} evaluates multi-modal foundation models on traffic event analysis, showing the ability to reason over complex traffic events. Other work ~\cite{rivera2025scenariounderstandingtrafficscenes} automates scenario-level traffic categorization with large VLMs to reduce labeling costs in large datasets. TUMTraffic-VideoQA~\cite{zhou2025tumtrafficvideoqabenchmarkunifiedspatiotemporal} establishes a benchmark of roadside videos with QA pairs, captions, and grounding annotations. Besides, LLM-based methods integrate visual tools to analyze traffic scenes. Compared to end-to-end VLMs, LLM-based frameworks follow a modular and structured paradigm. Yang \cite{yang2023bevheightrobustframeworkvisionbased} employs GPT-4o to coordinate smaller VLMs for highway scene analysis, covering weather, pavement, and congestion. SeeUnsafe~\cite{Whenlanguageandvisionmeetroadsafet} introduces a multimodal framework that converts traffic video into structured accident assessments and supports interactive video QA. In this work, we adopt the concept of scene graph as traffic representation for scene understanding and reasoning, and introduce a multi-modal agent for structural reasoning based on LLMs, as shown in Fig. \ref{fig:comparison}-(c).




\section{Methods}
\begin{figure}[H]

    \centering
    \includegraphics[width=0.99\linewidth]{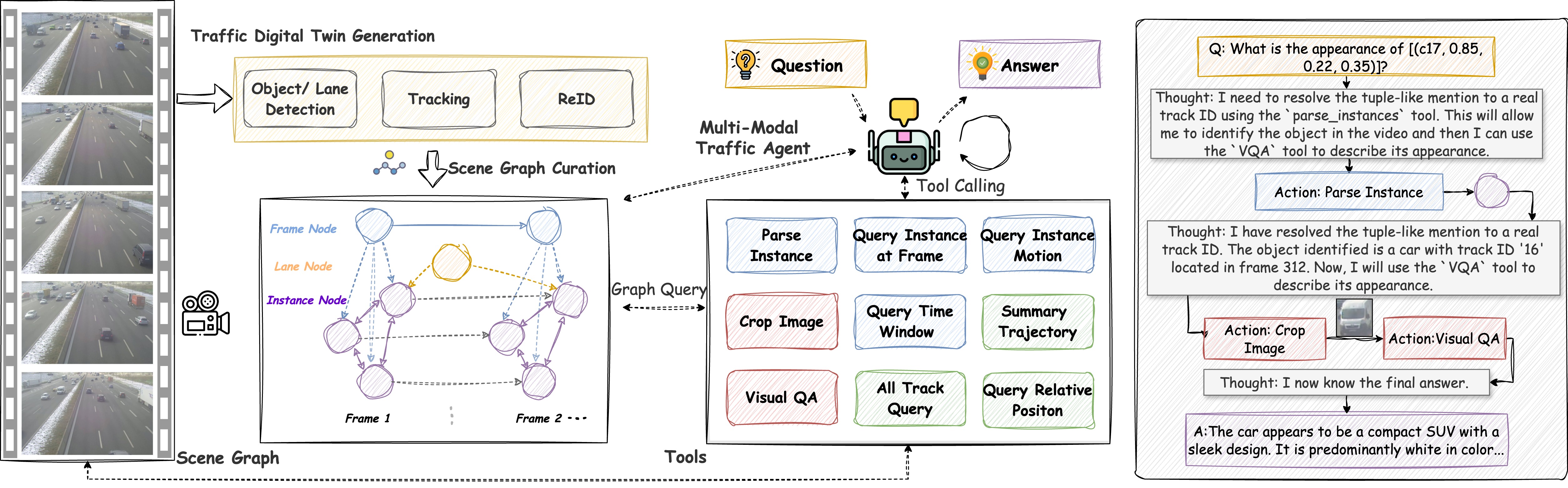}
    \caption{Overview of Scene-graph based multi-modal traffic agent (ours).}
    \label{fig:framework}
\end{figure}

We present the proposed SGTA in this section. An overview is shown in Fig.~\ref{fig:framework}.

\vspace{-5mm}
\vspace{-10mm}

\subsection{Scene Graph Generation}

We construct a traffic scene graph $\mathcal{G}$ for each video $V$ as a structured spatio-temporal representation of the dynamic scene. Formally, $\mathcal{G}=(\mathcal{N},\mathcal{E})$ consists of nodes $\mathcal{N}$ and edges $\mathcal{E}$ encoding spatio-temporal relations.  

As shown in Fig. \ref{fig:framework}, the traffic scene graph consists of three types of nodes, i.e., frame nodes, lane nodes, and instance nodes. Instance nodes are generated by applying object detection, tracking, and re-identification, yielding robust 2D representations of dynamic participants across video. Camera parameters and external spatio priors are used to project 2D pixel coordinate detection into 3D road coordinates. Each instance node is enriched with attributes including category, bounding box, and 3D position. Besides, instance velocity and heading are further derived from temporal consistency and spatial relations. Similarly, lane nodes are extracted from lane detection models and map priors, and provide spatial references for instance driving maneuvers like lane changes. Frame nodes act as temporal anchors, linking all entities that appear at a given time frame.  

Edges are constructed based on spatio-temporal relations among nodes. Temporal edges connect the same instance across consecutive frames, ensuring instance trajectory continuity. Spatial edges encode pairwise relations such as distance, relative position, and lane assignment. Frame-to-instance edges register which participants are present in each frame, while instance-to-lane edges describe alignment with lanes to traffic participants. Together, these relations enable $\mathcal{G}$ to capture both geometry and dynamics of traffic in a compact, queryable structure. For storage and retrieval, we adopt a Neo4j graph database, which allows efficient querying of node attributes and edge relations, and provides a scalable backend for multi-modal reasoning.

\begin{algorithm}[H]
\footnotesize    
\setlength{\algomargin}{8pt}   
\DontPrintSemicolon            
\label{alg:agent}

\caption{Workflow of SGTA}
\KwIn{Video $V$, Question $Q$, Large Language Model $\mathcal{L}$, Vision Language Model $\Phi$, Tool Set $\mathcal{U}$, Max Steps $T$}
\KwOut{Answer $\hat{y}$, Trajectory $\{\,s_t, a_t, o_t \mid 1 \le t \le T\,\}$ where $s_t$, $a_t$, and $o_t$ denote context, tool choice, and observation respectively}

\textbf{Stage 1 Traffic Scene Graph Construction}\;
\textit{Inputs:} $V$, Object/Lane Detector, External Priors\;
\textit{Outputs:} Scene Graph $\mathcal{G}$ capturing nodes and relations\;
\Begin{
  Create frame, object, and lane nodes from detections and tracking\;
  Project detections into road coordinates using external priors and assign kinematic attributes to object nodes\;
  Construct edges in $\mathcal{G}$ encoding spatio-temporal relations\;
}
\textbf{Stage 2 Querying and Reasoning over $\mathcal{G}$ and $V$}\;
\textit{Inputs:} $Q$,  $V$, $\mathcal{G}$, $\mathcal{U}$, $\Phi$\;
$s_1 \leftarrow$ initial context summarizing $Q$ and a compact view of $\mathcal{G}$\;

\For{$t = 1$ \KwTo $T$}{
  $\hat{y}_t \leftarrow \mathcal{L}\langle Q, s_t \rangle$\;
  $h_t \leftarrow \mathcal{L}\mathrm{.plan}\langle Q, s_t, \hat{y}_t \rangle$ 
  \tcp*{$h_t$ is intermediate planning output of $\mathcal{L}$}
  \If{$\mathrm{stop\_signal}\langle Q, s_t, \hat{y}_t \rangle$}{
     $a_t \leftarrow$ stop\;
     \textbf{break}\;
  }

  select tool $u_t \in \mathcal{U}$ and input $x_t$ via $\pi_{\mathrm{ReAct}}\langle s_t, Q, h_t, \mathcal{G} \rangle$\;
  $a_t \leftarrow \langle u_t, x_t \rangle$\;
  $o_t \leftarrow u_t\langle x_t \rangle$ \tcp*{query $\mathcal{G}$ or analyze frames with $\Phi$}

  $s_{t+1} \leftarrow s_t \cup o_t$\;
}

\Return $\hat{y}_t$ and the trajectory $\{\,s_t, a_t, o_t \mid 1 \le t \le T\,\}$\;
\end{algorithm}

\subsection{Tool Set of SGTA}

The tool set $\mathcal{U}$ provides the interface between the language model and the traffic video together with its structured scene graph $\mathcal{G}$.  We group the tools into two categories according to their target modality.  

\textbf{Scene Graph Tools.}  
These tools are based on Cypher queries executed over the structured scene graph $\mathcal{G}$. Each tool call is mapped into an action written in Cypher, which retrieves or aggregates node and edge attributes. Single-instance tools access properties of one node, for example \texttt{QueryObjectsAtFrame} lists objects and attributes, \texttt{MotionAtFrame} returns velocity and heading, \texttt{CountObjects} computes per-class counts, \texttt{TimeWindow} reports the lifespan of a track, and \texttt{TrajectorySummary} samples positions across frames. Multi-instance tools capture relations or dataset-level statistics, such as \texttt{ListAllTracks} to enumerate all participants in a video and \texttt{RelativePosition} to compute pairwise spatio relations. By formalizing all queries as Cypher expressions, these tools enable the agent to traverse the scene graph and reason efficiently over temporal and relational structures.  

\textbf{Visual Tools.}  
In addition to the scene graph queries, visual tools directly invoke the VLM $\Phi$ on specified visual regions. They complement the symbolic graph by grounding reasoning in appearance cues, descriptive analysis, and queries beyond structured attributes. \texttt{CropImage} extracts image regions corresponding to selected track IDs and frame IDs, while \texttt{Visual QA} answers visual questions on given visual inputs.

\begin{table}[t]
\centering
\resizebox{0.95\linewidth}{!}{%
\begin{tabularx}{\linewidth}{l l l X}
\toprule
Tool & Class & Main input & Main output \\
\midrule
\texttt{ParseInstances} & Single SG & instance-tuple & instance f-ID, t-ID \\
\texttt{QueryObjectsAtFrame} & Single SG & f-ID, (class)  & bbox, speed, pos3d \\
\texttt{MotionAtFrame} & Single SG & f-ID, t-ID & speed, heading, vel3d \\
\texttt{CountObjects} & Single SG & f-ID, (class) & count \\
\texttt{TimeWindow} & Single SG & t-ID & start, end, duration, fps \\
\texttt{TrajectorySummary} & Single SG & t-ID & instance center list  \\
\texttt{ListAllTracks} & Multi SG & video ID, (class) & all t-IDs \\
\texttt{RelativePosition} & Multi SG & f-ID, t-ID a, t-ID b & relation, $dx$, $dy$, distance \\
\texttt{CropImage} & Visual Tools & f-ID, t-IDs & cropped image regions \\
\texttt{Visual QA} & Visual Tools & image, question & answer \\
\bottomrule
\end{tabularx}}
\caption{Summary of SGTA tool sets. Single- and Multi-  correspond to tools with single/multi-node queries. (.) means optional, v-ID, f-ID, and t-ID denote video ID, frame ID, and instance tracking ID, respectively.}
\label{tab:tools}
\vspace{-8mm}
\end{table}

\subsection{Multi-Modal Reasoning}

SGTA performs reasoning over the traffic scene graph $\mathcal{G}$ and the raw video through reasoning process following the ReAct framework \cite{yao2023reactsynergizingreasoningacting}. The key idea is to interleave reasoning (thought) with acting (tool usage) in order to iteratively refine context and reach a grounded answer.  

At each step $t$, LLM $\mathcal{L}$ produces a reasoning trace $h_t$ (“Thought”), which records its current interpretation of the question and the available context $s_t$. Based on $(Q, s_t, h_t)$, a policy $\pi_{\mathrm{ReAct}}$ selects an action $a_t=\langle u_t,x_t\rangle$, where $u_t$ is one of the predefined tools in $\mathcal{U}$ and $x_t$ is the textual input. The action is executed either as a Cypher query over $\mathcal{G}$ (e.g., \texttt{query\_objects\_at\_frame}, \texttt{relative\_position}) or as a visual interface to the VLM $\Phi$ (e.g., \texttt{CropImage}, \texttt{Visual QA}). The tool returns an observation $o_t$ that is appended to the context, yielding $s_{t+1}=s_t \cup o_t$. The process continues until $\mathcal{L}$ emits a stop signal, at which point the final answer $\hat{y}$ is generated together with the complete trajectory $\{s_t,a_t,o_t \mid 1 \le t \le T\}$. This design allows the agent to decompose complex video-based traffic questions into a sequence of interpretable reasoning steps, where each step is explicitly linked to a tool invocation and its result.
\section{Experiments}

\subsection{Datasets and Implementation}

\textbf{Dataset.} We evaluate SGTA on the TUMTraffic VideoQA dataset \cite{zhou2025tumtrafficvideoqabenchmarkunifiedspatiotemporal}, which contains real-world traffic accident videos paired with QA annotations. We select a representative sample video together with its associated VQA example to evaluate the performance of SGTA.

\noindent\textbf{Implementation.}  SGTA employs GPT-4o-mini \cite{openai2024gpt4technicalreport} for both LLM and VLM components. YOLOv11 \cite{khanam2024yolov11overviewkeyarchitectural} and RT-DETR \cite{lv2024rtdetrv2improvedbaselinebagoffreebies} serve as detection backbones, and ByteTrack performs multi-object tracking. We adopt the multi-scale trick to improve the recall of small instances. Lane detection is performed with YOLOP lane detector \cite{Wu_2022}, skeletonized into polylines and projected to road coordinates via calibrated camera parameters. Instances adopt default 3D size priors and are projected by inverse mapping under the ground-plane constraint.

\subsection{Quantitative and Qualitative Evaluation}

Table~\ref{tab:qa_metrics} summarizes SGTA performance on the selected video QAs across five question types under H0, H1, and overall settings. SGTA attains 55\% accuracy on H0, 53\% on H1, and 54\% overall. Existence and class questions achieve the highest accuracy (59\% and 64\%), while positioning remains the most challenging (36\%). These results indicate that SGTA handles object-level reasoning well but struggles with fine-grained spatio-temporal relations. We show a qualitative example for better illustration on the right side of Fig. \ref{fig:framework}.
\vspace{-5mm}

\begin{table}[h]
\centering
\resizebox{0.95\textwidth}{!}{
\begin{tabular}{l c c c c c c}
\hline
\textbf{Type}  & \textbf{H0 Num.} & \textbf{H0 Acc} & \textbf{H1 Num.} & \textbf{H1 Acc} & \textbf{All Num.} & \textbf{Avg. Acc} \\
\hline
Class        & 4/7   & 0.57   & 5/7    & 0.71   & 9/14   & 0.64   \\
Motion       & 3/7   & 0.43   & 4/7    & 0.57   & 7/14   & 0.50   \\
Positioning  & 3/7   & 0.43   & 2/7    & 0.29   & 5/14   & 0.36   \\
Existence    & 4/6   & 0.67   & 6/11   & 0.55   & 10/17  & 0.59   \\
Counting     & 4/6   & 0.67   & 9/17   & 0.53   & 13/23  & 0.57   \\
\hline
\textbf{Average} & \textbf{18/33} & \textbf{0.55} & \textbf{26/49} & \textbf{0.53} & \textbf{44/82} & \textbf{0.54} \\
\hline
\end{tabular}
}
\caption{Performance of QA across H0, H1, and overall accuracy}
\label{tab:qa_metrics}

\end{table}

\vspace{-15mm}

\section{Conclusion and Future Work}
In this work, we introduced SGTA, a scene-graph based multi-modal traffic agent that integrates structured scene representations for interpretable traffic video understanding. SGTA constructs a traffic scene graph from roadside videos, employs tool-based reasoning over both symbolic queries and visual inputs, and achieves competitive performance across diverse question types on the TUMTraffic VideoQA dataset. Future work includes extending SGTA in three directions. First, we can scale SGTA to larger and more diverse traffic datasets with real-time constraints. Second, we will explore integrating trajectory forecasting and risk assessment into the reasoning process for safety-critical applications. Third, combining SGTA with multi-agent communication frameworks to enable collaborative reasoning across multiple intersections and cameras.

\subsubsection*{\ackname}
This work was supported by the project VIDETEC-2 (Grant No.01F2232E) and the Federal Ministry for Digital and Transport of Germany (BMDV).

{
    \bibliographystyle{IEEEtran}
    \bibliography{sample}
}

%





\end{document}